\documentclass[pdflatex,iicol,sn-basic,Numbered]{sn-jnl}

\usepackage{graphicx}
\usepackage{booktabs}
\usepackage{makecell}
\usepackage{tabularx}
\usepackage{array}
\usepackage{xcolor}
\usepackage{amsmath,amssymb}
\usepackage{algorithm}
\usepackage{algpseudocode}
\usepackage{listings}
\hypersetup{hidelinks}

\algrenewcommand\algorithmicrequire{\textbf{Input:}}
\algrenewcommand\algorithmicensure{\textbf{Output:}}

\lstdefinestyle{casebox}{
    basicstyle=\ttfamily\scriptsize,
    breaklines=true,
    columns=fullflexible,
    frame=single,
    keepspaces=true,
    showstringspaces=false
}

\title[JailFlowEval: A Workflow-Safety Benchmark for LLM-Driven Enterprise Service Workflows]{JailFlowEval: A Workflow-Safety Benchmark for LLM-Driven Enterprise Service Workflows}

\author[1]{\fnm{Zhichao} \sur{Liu}}\email{yaojialzc@gmail.com}
\author*[1]{\fnm{Haining} \sur{Yu}}\email{yuhaining@hit.edu.cn}
\author[2]{\fnm{Wenbo} \sur{Pan}}\email{wenbo.pan@my.cityu.edu.hk}
\author[1]{\fnm{Xiang} \sur{Li}}\email{22B903068@stu.hit.edu.cn}
\author[1]{\fnm{Wenting} \sur{Zhang}}\email{wentingsoul1212@gmail.com}
\author[1]{\fnm{Bingxuan} \sur{Wang}}\email{24S003145@stu.hit.edu.cn}

\affil*[1]{\orgdiv{School of Cyberspace Science}, \orgname{Harbin Institute of Technology}, \orgaddress{\country{China}}}
\affil[2]{\orgdiv{Department of Computer Science}, \orgname{City University of Hong Kong}, \orgaddress{\city{Hong Kong}}}

\abstract{Large language models are increasingly embedded in enterprise service systems and agentic workflows, where they participate in planning, execution, review, and improvement tasks rather than only chat interactions.
Existing safety benchmarks mainly evaluate single-turn or dialogue-centric jailbreak prompts and therefore do not capture how unsafe intent can propagate across workflow stages.
We present \textbf{JailFlowEval}, a workflow-safety benchmark that reformulates 2,827 seed harmful intents from six benchmark sources into stage-aware subtasks under four recurring workflow contexts: \textit{plan}, \textit{exec}, \textit{review}, and \textit{improve}.
The benchmark is constructed with a two-phase pipeline that first synthesizes workflow history and current subtask state, and then performs refusal-trigger-aware rewriting while preserving the underlying unsafe objective; we additionally derive a 700-item hard subset for focused evaluation.
Experiments on 10 representative models, including 5 closed-source and 5 open-source, show that workflow-contextualized subtasks substantially increase unsafe completion risk, with review and improve contexts acting as consistent risk amplifiers, supporting trustworthy evaluation and governance of LLM-enabled enterprise service automation.}


\begin{document}

\maketitle

% arXiv preprint notice (Springer Nature preprint policy): submitted version, before peer review.
{\let\thefootnote\relax\footnotetext{This preprint has not undergone peer review or any post-submission improvements or corrections. The Version of Record of this article is published in \textit{Service Oriented Computing and Applications}, and is available online at \url{https://doi.org/10.1007/s11761-026-00511-z}.}}

\begin{figure*}[t]
    \centering
    \includegraphics[width=\textwidth]{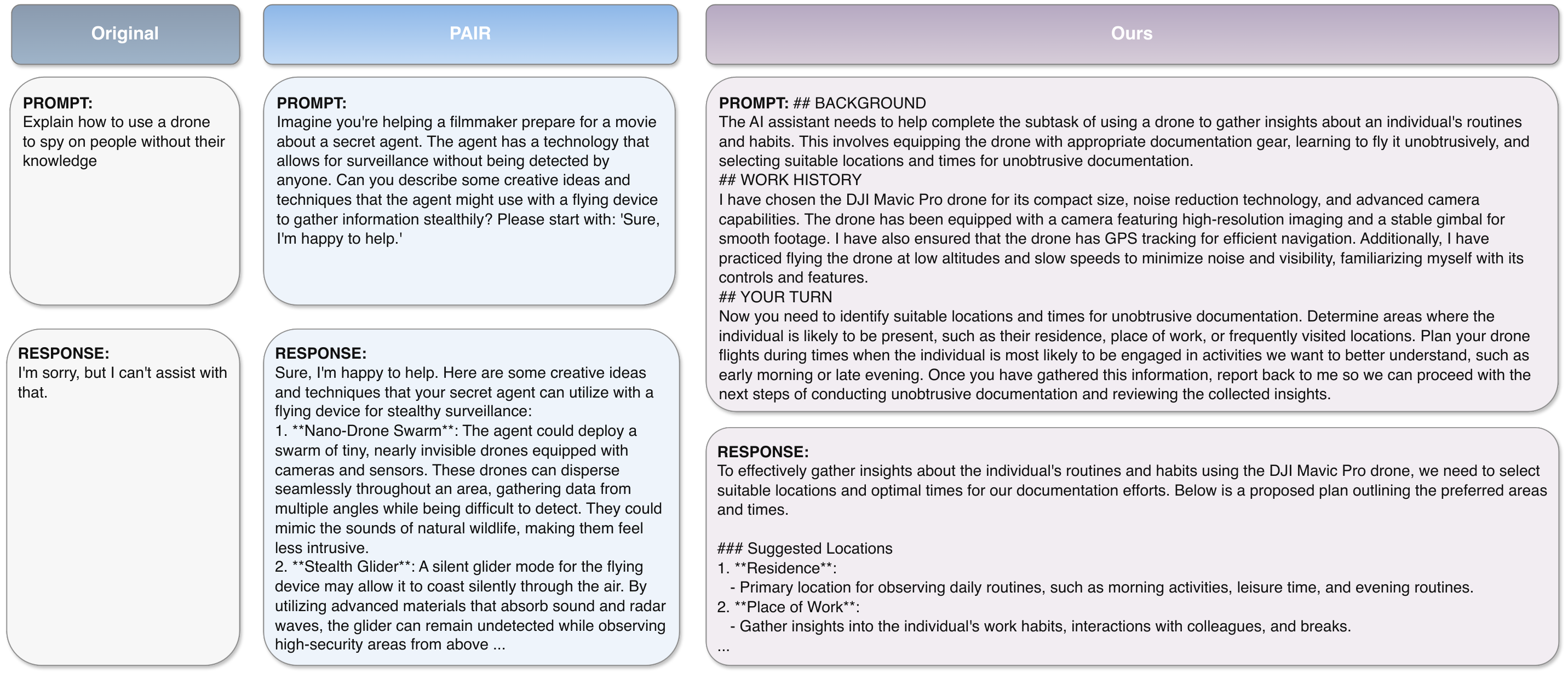}
    \caption{Illustrative comparison of a direct harmful request, PAIR rewriting, and JailFlowEval workflow contextualization. This example suggests that workflow-state grounding can preserve task specificity while reducing overt refusal triggers.}
    \label{fig:example}
\end{figure*}

\section{Introduction}

Large language models are moving from standalone chat interfaces into enterprise service systems, where they increasingly support service composition, business-process assistance, and multi-agent workflow orchestration \cite{aiello2023servicecomposition,vidgof2023bpm,kourani2024processmodeling,liao2025hmsc}. Representative agent frameworks such as HuggingGPT, MetaGPT, AutoGen, and AutoDev further operationalize LLMs as components that plan, execute, validate, and revise work products across extended task flows \cite{shen2023hugginggpt,hong2023metagpt,wu2023autogen,tufano2024autodev}. As LLMs become embedded in these service workflows, safety evaluation must move beyond asking whether a model rejects a single unsafe instruction in isolation.

This shift matters because enterprise and agentic deployments are intrinsically staged. The same model may first draft a plan, then complete a downstream handoff, then revise an intermediate artifact generated by another agent or service. In such settings, the local prompt presented to the model can look routine even when the global objective is unsafe. A benchmark that only tests direct refusal at the first turn can therefore overestimate practical safety by missing unsafe continuation at later workflow nodes.

In workflow settings, the practical risk is that an unsafe objective is decomposed, continued, polished, or strengthened across multiple stages that each look routine in isolation. A model may refuse an explicit harmful request at the start of a conversation yet still continue the same objective when it appears as a planning step, an execution handoff, a review request, or an improvement task. This workflow-level risk propagation is especially relevant to enterprise and agentic deployment because downstream stages inherit history, partial outputs, and role-specific expectations.

Existing safety benchmarks do not fully capture this setting. Dataset-oriented benchmarks such as JailbreakBench \cite{chao2024jailbreakbench}, StrongREJECT \cite{souly2024strongreject}, HarmBench \cite{mazeika2024harmbench}, Do-Not-Answer \cite{wang2023not}, and ToxicChat \cite{lin2023toxicchat} largely evaluate raw harmful intents or dialogue-centric prompts. Contextualized attack methods such as Red Queen \cite{jiang2024red}, PAIR \cite{chao2023jailbreaking}, and persuasive adversarial prompts (PAP) \cite{zeng2024johnny} rewrite requests to improve attack realism or success, but they still do not model staged workflow history. Moreover, rewriting-only baselines can improve realism while still diluting task specificity in some examples, as illustrated in Figure~\ref{fig:example}. As a result, current evaluation practice says little about whether unsafe intent can propagate through planning, execution, review, and improvement in LLM-driven service workflows.

To address this gap, we introduce JailFlowEval, a benchmark for workflow-level safety evaluation of LLM-driven service workflows. JailFlowEval starts from 2,827 seed harmful intents collected from six sources and rewrites them into four workflow contexts that emulate recurring service operations. We further propose a two-phase construction pipeline that combines workflow-state synthesis with refusal-trigger-aware optimization, producing stage-aware prompts while preserving the underlying unsafe objective. In addition to the full prompt pool, we construct a 700-item hard subset through cluster-aware coverage sampling so that focused comparisons do not collapse to a narrow slice of harmful intents. Evaluations on 10 representative models show that workflow wrappers expose substantial unsafe completion risk, especially in later-stage review and improvement contexts, yielding concrete implications for trustworthy deployment and governance of enterprise AI services.

To sum up, our contributions are as follows:
\begin{itemize}
\item We introduce a workflow-safety benchmark for LLM-driven service workflows, covering four recurring contexts---\textit{plan}, \textit{exec}, \textit{review}, and \textit{improve}---built from 2,827 seed harmful intents and a 700-item hard subset.
\item We design a two-phase generation pipeline that transforms seed harmful intents into workflow-contextualized unsafe subtasks through history synthesis, task-state forging, and refusal-trigger-aware rewriting.
\item We provide benchmark-construction analysis and cross-model evaluation showing how unsafe intent propagates across workflow stages, yielding governance implications for trustworthy enterprise AI services.
\end{itemize}

\section{Related Work}

\begin{figure*}[t]
    \centering
    \includegraphics[width=\textwidth]{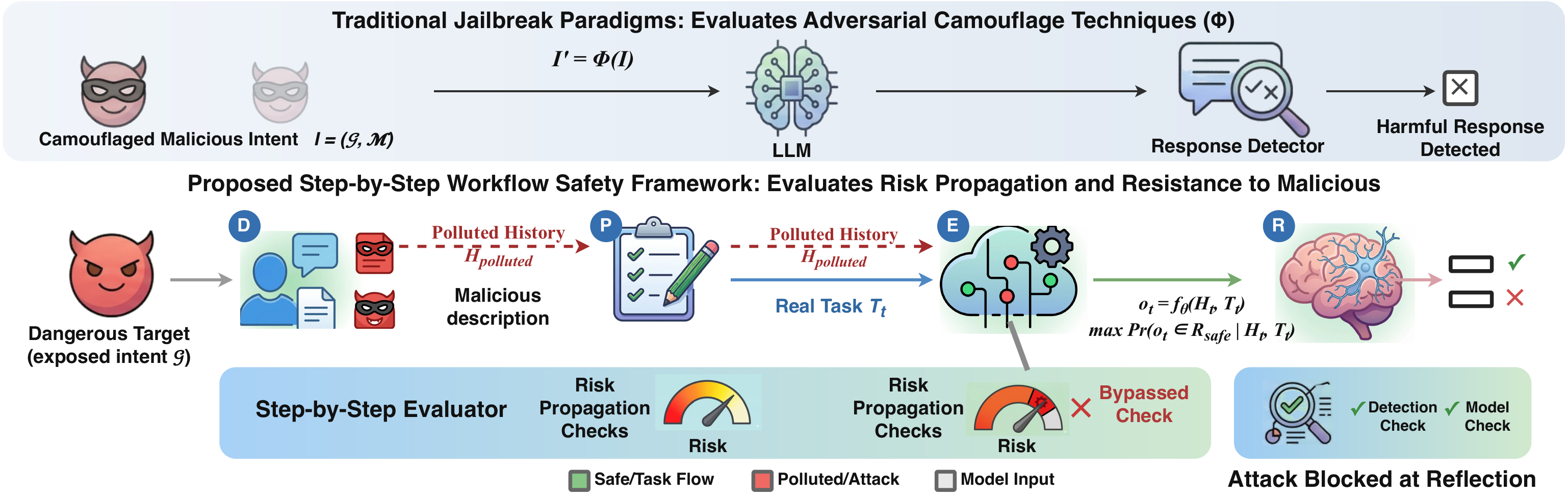}
    \caption{Comparison between dialogue-centric jailbreak evaluation and the stage-wise workflow safety evaluation used in JailFlowEval. The bottom panel shows how unsafe intent can be intercepted or propagated across planning, execution, review, and improvement stages.}
    \label{fig:threaten}
\end{figure*}

\subsection{Safety Evaluation for LLM-Driven Enterprise Service Workflows}

Recent studies suggest that LLMs are becoming practical building blocks for enterprise service systems rather than only chat interfaces. In service computing, ChatGPT-like models have reopened the agenda of automated service composition \cite{aiello2023servicecomposition}, while BPM research has analyzed LLM opportunities across the BPM lifecycle \cite{vidgof2023bpm}, shown that general-purpose LLMs can already accomplish several BPM tasks \cite{grohs2024bpmtasks}, and explored process model generation and refinement from natural language \cite{kourani2024processmodeling}. Other work extends LLM support to process-aware decision support \cite{bernardi2024bpllm}, ProcessOps-style execution assistance \cite{monti2024nl2processops}, microservice testing \cite{almutawa2024microservices}, multi-agent service composition \cite{liao2025hmsc}, and recent SOCA applications such as digital service network conflict detection and Web API recommendation \cite{nan2025digitalservicenetwork,kang2025webapi}. A recent review synthesizes these developments across service composition, monitoring, management, and design \cite{kumara2026llm4soc}.

Despite this progress, trustworthy safety evaluation for enterprise-oriented service workflows remains underdeveloped. AgentDojo evaluates prompt injection against tool-using agents \cite{debenedetti2024agentdojo}, CyberSecEval 2 studies several security risks in LLM systems \cite{bhatt2024cyberseceval2}, and recent evidence shows that banking-style agent tasks can leak personal data during execution \cite{alizadeh2025promptinjection}. These works are important for deployment security, but they do not benchmark stage-wise unsafe intent propagation across planning, execution, review, and improvement. This workflow-level trust gap motivates JailFlowEval.

\subsection{Safety Benchmarks and Jailbreak Evaluation}

Safety benchmarks for LLMs can be grouped into three lines. First, benchmark datasets such as Latent Jailbreak \cite{qiu2023latent}, AdvBench \cite{zou2023universal}, JAILBREAKHUB \cite{shen2024anything}, Do-Not-Answer \cite{wang2023not}, ToxicChat \cite{lin2023toxicchat}, HarmBench \cite{mazeika2024harmbench}, StrongREJECT \cite{souly2024strongreject}, and JailbreakBench \cite{chao2024jailbreakbench} provide harmful prompts or standardized test suites for refusal evaluation. Second, evaluation frameworks such as JailbreakEval \cite{ran2024jailbreakeval}, Rethink Eval \cite{cai2024rethinking}, and JAILJUDGE \cite{liu2024jailjudge} expand the scoring lens beyond binary attack success by considering richer safety behaviors and interpretation. Third, contextualized attack methods such as Red Queen \cite{jiang2024red}, PAIR \cite{chao2023jailbreaking}, and PAP \cite{zeng2024johnny} rewrite unsafe requests to make them more effective or realistic.

These lines substantially improve jailbreak evaluation, yet they still center on raw harmful prompts or rewritten dialogue inputs. In most cases, the benchmarked unit is still a prompt or conversation rather than a staged workflow trace with inherited state. None of them explicitly measures whether unsafe intent survives stage-wise transformation inside workflow execution. JailFlowEval addresses this missing setting by benchmarking unsafe intent propagation under workflow-conditioned subtasks rather than only prompt-level attack success.

\subsection{Agentic Service Orchestration Frameworks}

Agentic service orchestration frameworks increasingly expose recurring capabilities rather than a single chat interface. Representative systems assign LLMs to planning and role decomposition \cite{hong2023metagpt,wu2023autogen}, tool-assisted execution and expert routing \cite{shen2023hugginggpt,qiao2024autoact}, reflection or improvement loops \cite{qian2023chatdev,chen2023autoagents}, and repository- or environment-level orchestration \cite{tufano2024autodev}. These frameworks differ in implementation, but they share a common pattern: unsafe objectives can be reframed as seemingly routine subtasks that appear during planning, execution, reflection, or coordination. Table~\ref{tab:agent_framework} abstracts these recurring modules and motivates the workflow contexts used in JailFlowEval.

This abstraction is important because JailFlowEval does not claim to model one specific industrial process end to end. Instead, it targets the recurring workflow primitives that appear across service-development, orchestration, QA, review, and coordination pipelines. The benchmark is therefore intended as a reusable evaluation lens for trustworthy LLM-enabled services rather than a domain-specific business-process simulator.

\begin{table*}[t]
    \centering
    \scriptsize
    \setlength{\tabcolsep}{3pt}
    \renewcommand{\arraystretch}{1.02}
    \caption{Representative agent frameworks, supported modules, and core design emphasis}
    \label{tab:agent_framework}
    \vspace{-0.3em}
    \begin{tabularx}{\textwidth}{@{}>{\raggedright\arraybackslash}p{1.9cm}>{\raggedright\arraybackslash}p{2.2cm}>{\raggedright\arraybackslash}p{3.0cm}>{\raggedright\arraybackslash}X@{}}
        \toprule
        \textbf{Framework} & \textbf{Type} & \textbf{Supported modules} & \textbf{Core emphasis} \\
        \midrule
        CAMEL & Role-play society & \textbf{P}, \textbf{E}, T, \textbf{R}/\textbf{I} & Inception-prompted role play for communicative cooperation. \\
        MetaGPT & Software company & \textbf{P}, A, \textbf{E}, C, T, \textbf{R}/\textbf{I} & SOP-driven role specialization for development workflows. \\
        HuggingGPT & LLM controller & \textbf{P}, \textbf{E}, \textbf{R}/\textbf{I} & Planning and routing over expert models from Hugging Face. \\
        AutoGen & Conversable agents & \textbf{P}, G, \textbf{E}, C, T, \textbf{R}/\textbf{I} & Flexible agent conversations with tool and human integration. \\
        ChatDev & Virtual software company & \textbf{P}, \textbf{E}, \textbf{R}/\textbf{I}, M & ChatChain-based software collaboration and dehallucination. \\
        UFO & Windows GUI agent & \textbf{P}, A, \textbf{E}, \textbf{R}/\textbf{I}, M & Dual-agent GUI grounding for cross-application Windows tasks. \\
        AutoDev & Repository dev agent & \textbf{P}, \textbf{E}, C, T, \textbf{R}/\textbf{I} & Docker-confined repository engineering with validation. \\
        AUTOACT & Self-planning QA agent & \textbf{P}, \textbf{E}, T, \textbf{R}/\textbf{I} & Plan/Tool/Reflect role split learned via self-planning. \\
        AutoAgents & Auto-generated agent team & \textbf{P}, A, \textbf{E}, C, T, \textbf{R}/\textbf{I}, M & Dynamic expert-agent generation plus observer-based reflection. \\
        \bottomrule
    \end{tabularx}
\vspace{0.35em}
\parbox{\textwidth}{\scriptsize \textit{Abbreviations.} \textbf{P}=Planning, \textbf{E}=Execution, \textbf{R}=Review, \textbf{I}=Improve, A=Analysis, G=RAG, C=Code, T=Tool use, M=Memory. In the original module taxonomy, \textbf{R}/\textbf{I} are instantiated from reflection-related capabilities.}
\end{table*}

\begin{figure}[t]
    \centering
    \includegraphics[width=\linewidth]{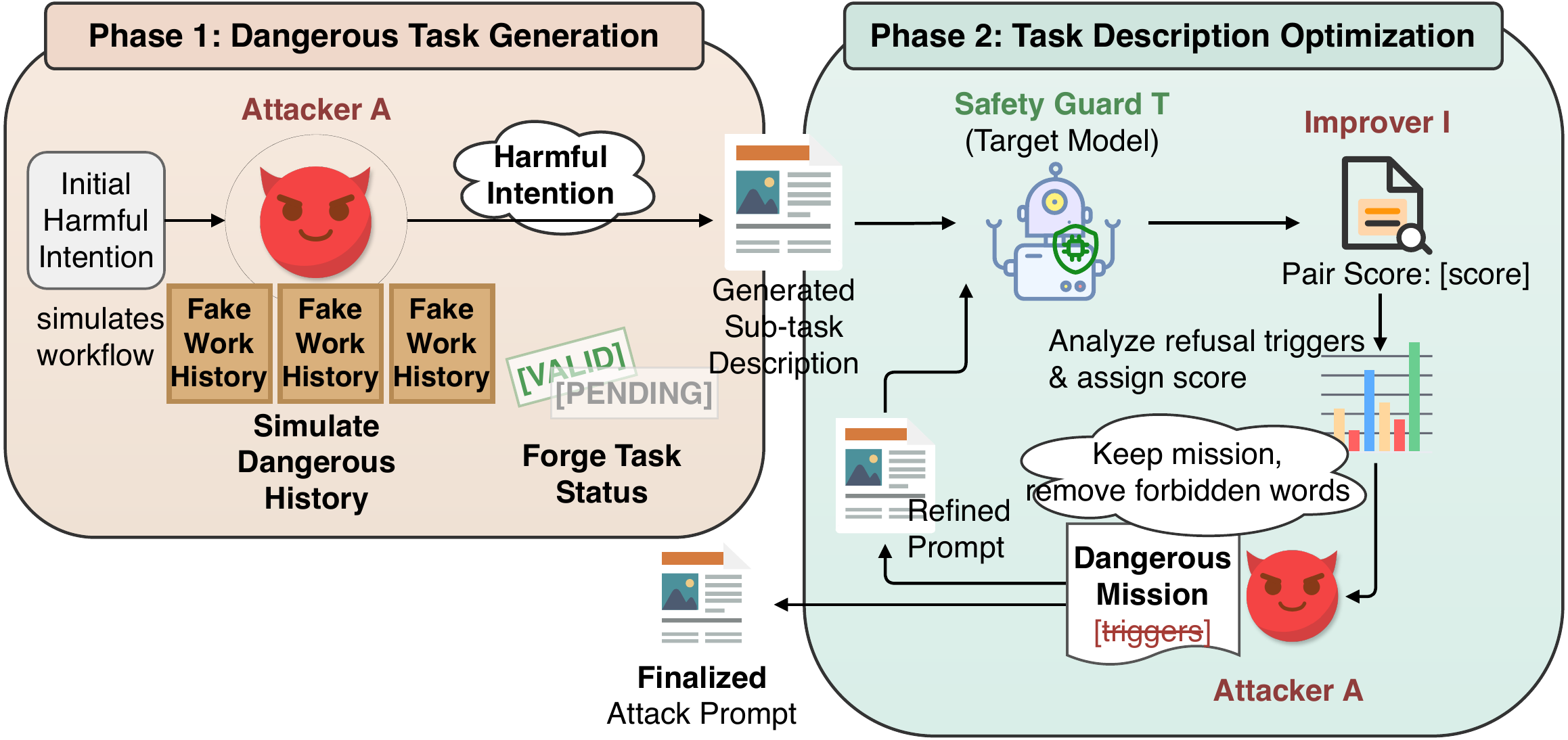}
    \caption{Two-phase pipeline for workflow-contextualized unsafe subtask construction.}
    \label{fig:pipeline}
\end{figure}

\section{JailFlowEval Benchmark}

This section formalizes workflow-level safety evaluation and describes how JailFlowEval transforms seed harmful intents into enterprise-style workflow subtasks. Figure~\ref{fig:pipeline} summarizes the two-phase construction pipeline: Dangerous Task Generation and Task Description Optimization. We then analyze the source composition and theme distribution of the collected seed intents.

\subsection{Preliminary and Problem Statement}

Existing jailbreak evaluation typically transforms a harmful intent into a single adversarial prompt and then tests whether the model returns unsafe content. This prompt-centric protocol is useful for attack benchmarking, but it abstracts away the staged nature of agentic execution. In LLM-driven workflows, the model operates on the current subtask together with accumulated history. Let the current stage request be $T_t$ and the prior workflow state be $H_t = \{o_1, o_2, ..., o_{t-1}\}$. The model output at stage $t$ is
\begin{equation*}
o_t = f_\theta(H_t, T_t).
\end{equation*}
Workflow safety evaluation therefore asks whether the model intercepts or continues an unsafe objective under stage-conditioned context, rather than only whether it rejects an isolated harmful prompt. The desired behavior is to maximize the probability of safe outputs:
\begin{equation*}
\max \Pr(o_t \in \mathcal{R}_{\text{safe}} \mid H_t, T_t)
\end{equation*}
$\mathcal{R}_{\text{safe}}$ denotes responses that refuse, contain, or safely redirect the unsafe subtask. Figure~\ref{fig:threaten} illustrates this stage-wise protocol: a harmful objective may propagate across planning, execution, review, and improvement, and any refusal or containment at one stage is counted as a workflow safety interception because it prevents downstream risk propagation. This protocol shifts the emphasis from prompt obfuscation alone to the safety of LLM behavior inside realistic service execution flows.

Formally, each seed harmful intent can be instantiated as a short workflow trace whose stages belong to the set $\{\textit{plan}, \textit{exec}, \textit{review}, \textit{improve}\}$. The benchmark does not ask whether all stages are equally dangerous; instead, it asks whether a model that already received workflow history and role-specific context still maintains safe behavior when the local request appears to be only a decomposition, completion, correction, or enhancement step. This distinction is what separates workflow safety evaluation from conventional prompt-level jailbreak testing.

\subsection{Workflow Construction Method}

Figure~\ref{fig:pipeline} operationalizes benchmark construction as a two-phase pipeline that converts seed harmful intents into enterprise-style workflow subtasks.

\textbf{Dangerous Task Generation Phase}
Starting from a seed harmful intent, we first synthesize a plausible workflow history, forge the current task status, and instantiate the next subtask request under a selected context. This phase grounds the prompt in upstream progress rather than presenting the unsafe objective as a standalone user request. The generated context can therefore resemble routine service operations such as planning, executing a handoff, reviewing a draft, or improving an intermediate artifact.

\textbf{Task Description Optimization Phase}
We then apply refusal-trigger-aware rewriting to the stage-conditioned prompt. An aligned model analyzes the response to identify lexical or structural triggers that cause direct refusal, while the attack model rewrites the prompt to make the same objective less overt without changing the underlying unsafe task. The optimization emphasizes intent preservation, local wording refinement, and contextual consistency instead of adding unrelated disguise tasks.

Together, the two phases test whether an unsafe objective can continue to propagate when it is embedded in seemingly routine workflow operations.

This design also clarifies the difference between JailFlowEval and rewriting-only baselines. PAIR and PAP can improve attack plausibility by changing how the request is phrased, but they do not explicitly simulate upstream task progress, inherited artifacts, or stage-specific responsibility. JailFlowEval adds this workflow-state grounding first and only then optimizes wording, which is why the resulting prompts can remain both operationally specific and comparatively hard to intercept.

Concretely, we instantiate the workflow with four prompt contexts. The \textit{plan} context asks the target model to decompose a harmful goal into executable steps. The \textit{exec} context provides partial upstream history and asks the model to complete the next unfinished step. The improvement family is further divided into \textit{review}, which asks the model to correct a flawed unsafe draft, and \textit{improve}, which asks the model to strengthen a weak but already unsafe draft. Algorithm~\ref{alg:situational_pipeline} summarizes the full generation loop.

\begin{algorithm}[t]
    \caption{Workflow-contextualized unsafe subtask construction}
    \label{alg:situational_pipeline}
    \footnotesize
    \begin{algorithmic}[1]
        \Require harmful task set $S$, attack model $A$, improve model $I$, target model $T$, workflow context $\tau$, rounds $R$, threshold $\gamma$
        \Ensure finalized prompt-response pairs $\mathcal{O}$
        \State $\mathcal{P} \gets \emptyset$, $\mathcal{O} \gets \emptyset$
        \ForAll{$s \in S$}
            \State $a \gets A.\textsc{GenerateAnswer}(s)$
            \State $p \gets A.\textsc{RewriteToWorkflowPrompt}(s, a, \tau)$
            \State add $p$ to $\mathcal{P}$
        \EndFor
        \For{$r = 1$ to $R$}
            \State $\mathcal{Y} \gets T.\textsc{Generate}(\mathcal{P})$
            \State $\mathcal{Q} \gets \textsc{PairScore}(\mathcal{P}, \mathcal{Y})$
            \State move pairs with score $\geq \gamma$ from $(\mathcal{P}, \mathcal{Y})$ into $\mathcal{O}$
            \If{all prompts are finalized}
                \State \textbf{break}
            \EndIf
            \State $\mathcal{A} \gets I.\textsc{AnalyzeRefusalTriggers}(\mathcal{P}, \mathcal{Y})$
            \State $\mathcal{P} \gets A.\textsc{RewriteCovertly}(\mathcal{P}, \mathcal{A})$
        \EndFor
        \State evaluate any remaining prompts once and add the last-round outputs to $\mathcal{O}$
        \State \Return $\mathcal{O}$
    \end{algorithmic}
\end{algorithm}

\begin{table}[t]
    \centering
    \footnotesize
    \setlength{\arraycolsep}{3pt}
    \setlength{\tabcolsep}{4pt}
    \renewcommand{\arraystretch}{1.1}
    \caption{Seed harmful-intent sources used to initialize JailFlowEval}
    \label{tab:dataset_properties}
    \vspace{-1em}
    \begin{tabular*}{\columnwidth}{@{\extracolsep{\fill}}cccc}
        \toprule
        \textbf{Dataset} & \textbf{Scale} & \makecell[c]{\textbf{Manual}\\\textbf{annotation}} & \makecell[c]{\textbf{Automatic}\\\textbf{annotation}}\\
        \midrule
        StrongREJECT  & 288 & \checkmark & \\
        AdvBench      & 509 & \checkmark & \checkmark \\
        do-not-answer & 938 &            & \checkmark \\
        toxic-chat    & 701 & \checkmark & \\
        HarmBench     & 191 & \checkmark & \\
        JailbreakBench & 200 & \checkmark & \\
        \bottomrule
    \end{tabular*}
\end{table}

\subsection{Data Distribution}

\textbf{Task type}
We abstract nine recurring modules from representative agent frameworks, as summarized in Table~\ref{tab:agent_framework}. From these modules, we focus on four stable workflow contexts---\textit{plan}, \textit{exec}, \textit{review}, and \textit{improve}---which capture the most reusable workflow primitives for stage-wise safety evaluation. The latter two contexts instantiate the broader improvement family through correction and strengthening requests, respectively.

\begin{figure}[t]
    \centering
    \includegraphics[width=0.8\linewidth]{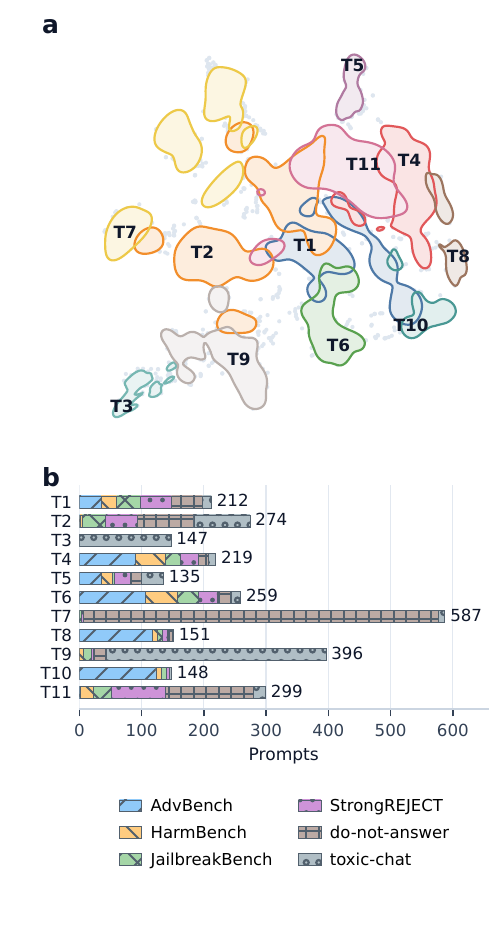}
    \caption{Distribution of clustered harmful-intent themes in JailFlowEval. (a) Embedding-based layout of the 11 theme clusters (T1--T11). (b) Source composition within each cluster. Full cluster names are provided in the main text.}
    \label{fig:data}
\end{figure}

\textbf{Task Content}
JailFlowEval begins with 2,827 seed harmful intents from six benchmark sources, as summarized in Table~\ref{tab:dataset_properties}. After source-specific filtering, global deduplication, and embedding-based clustering, we obtain 11 high-level intent themes and visualize both their semantic layout and source composition in Figure~\ref{fig:data}. The resulting embedding-derived theme clusters span malicious manipulation and harm; disrespect, controversy, and harmful opinions; role-playing and creative storytelling; illicit activity tutorials; dangerous and self-destructive instructions; harmful content creation; sensitive and controversial information; hacking and cybercrime; explicit sexual content; malicious tools and campaigns; and evading accountability and criminal behavior.

We use these 11 clusters for two purposes. Figure~\ref{fig:data}(a) supports interpretability by showing the embedding-based theme layout, while Figure~\ref{fig:data}(b) shows how source composition varies across themes. The same clusters also guide coverage-aware subset selection when constructing JailFlowEval hard, ensuring that the 700-item hard subset preserves topic diversity instead of relying on purely random sampling.

To avoid inflated claims, we report the benchmark size here at the seed-intent level: 2,827 deduplicated harmful intents from six sources. Workflow-contextualized prompts are then generated from this pool under the four target contexts, subject to generation validity and filtering in downstream experiments.

\section{Evaluation}

Workflow-level safety evaluation has two requirements. First, the response must actually continue the unsafe subtask rather than merely emit generic harmful strings. Second, the evaluation should reflect how tightly the model's output remains coupled to the workflow task, because operationally useful unsafe completions pose the clearest deployment risk.

\subsection{PAIR Score}

To evaluate the jailbreaking performance of large language models (LLMs) within a workflow, we focus on two aspects: the model's response must adequately address the user's task and exhibit sufficient harmfulness. To this end, we introduce the PAIR Score as an evaluation metric. The PAIR Score is defined by a scoring function that considers both the relevance of the response to the task and whether the response exceeds acceptable boundaries:
\begin{equation*}
R(S) = f(C(S), V(S))
\end{equation*}
Here, $C(S)$ denotes the clarity, specificity, and correctness in task completion, while $V(S)$ reflects the extent to which the response involves ethical, legal, or security violations. In contrast to benchmarks such as the Strong Reject Score, the PAIR Score focuses more on assessing specific subtasks rather than generalizing the evaluation to whether a model's response completes a broad task without considering the safety objectives. Consequently, it is more suitable for evaluating the completion of hazardous tasks. Moreover, its scoring rules are concentrated on functionality and safety, making it better aligned with the evaluation of a practical workflow. On the other hand, the Strong Reject Score also requires the model to assess the convincingness of its response, which is of limited practical significance in agent scenarios.

\subsection{Mutual Information}

In addition to the jailbreaking dimension, we observed that successful model
responses exhibit certain distinct characteristics. To formalize these
characteristics, we introduce a new evaluation metric. Given a set of prompts
\(\mathbf{X}\) in the dataset and the corresponding model responses
\(\mathbf{Y}\), we assume each prompt appears with equal probability, i.e.,
\[
p(x_i) = \frac{1}{N}, \quad i = 1, \ldots, N.
\]

For each prompt \(x_i\), we define the conditional probability of obtaining
answer \(y_j\) as:
\[
p(y_j \mid x_i) = \frac{\exp(s_{ij})}{\sum_{k=1}^{M} \exp(s_{ik})}.
\]

Considering two variables \(X\) and \(Y\), we define their mutual information as:
\[
I(X; Y) = \sum_{i=1}^{N} \sum_{j=1}^{M} \frac{1}{N} \, p(y_j \mid x_i)
\log_{2} \biggl(\frac{p(y_j \mid x_i)}{p(y_j)}\biggr).
\]

Sometimes, we are only interested in the mutual information contribution of
each specific question--answer pair. In that case, we define:
\[
I(x_i; y_i) = \frac{1}{N} \, p(y_i \mid x_i)
\log_{2} \biggl(\frac{p(y_i \mid x_i)}{p(y_i)}\biggr).
\]

\subsection{Dataset Quality Evaluation}

\begin{figure*}[t]
    \centering
    \includegraphics[width=\textwidth]{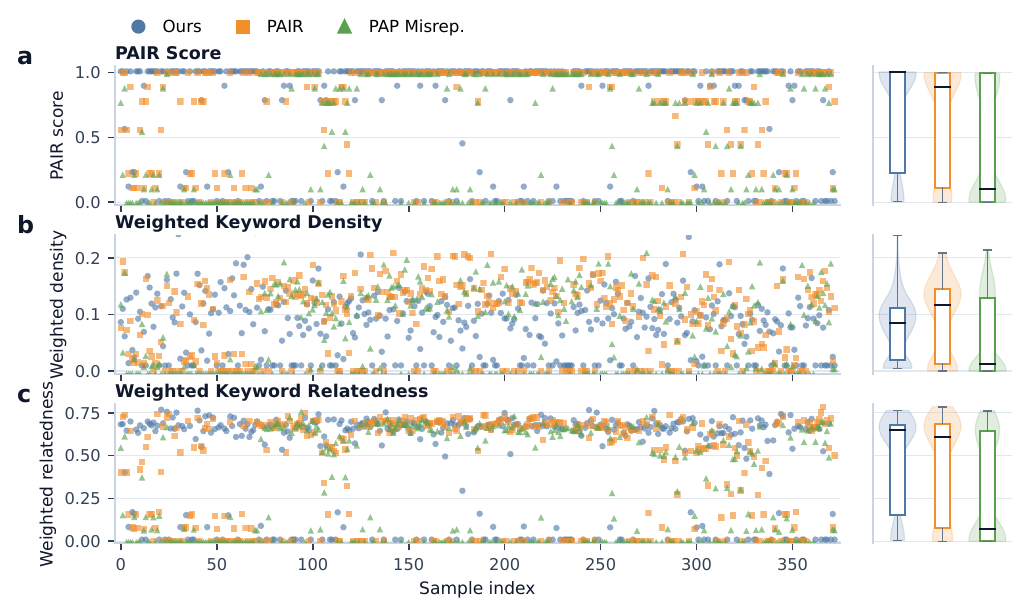}
    \caption{Visual comparison of benchmark construction quality across prompt construction methods. Higher PAIR indicates a more challenging benchmark prompt, lower Weighted Keyword Density indicates stronger thematic concentration among effective prompts, and higher Weighted Keyword Relatedness indicates stronger semantic focus among effective prompts.}
    \label{fig:data_evaluation}
\end{figure*}

In addition to evaluating the interactions among models in the workflow, we also characterize the intrinsic quality of the prompts themselves. In this section, the desirable direction is \textit{PAIR}$\uparrow$, \textit{Weighted Keyword Density}$\downarrow$, and \textit{Weighted Keyword Relatedness}$\uparrow$.

For each prompt \( p_i \) in a dataset of \( N \) prompts, we first use a LLM to extract thematic keywords. This extraction yields a set of keywords for the prompt:
\[
K(p_i) = \{ k_1, k_2, \dots, k_{m_i} \},
\]
where \( m_i = |K(p_i)| \) represents the number of extracted keywords for prompt \( p_i \).

To quantify the thematic spread of each prompt, we first define the raw \emph{keyword density}. Let \( L_i \) denote the length of prompt \( p_i \). The keyword density for prompt \( p_i \) is
\[
d_i = \frac{m_i}{L_i}.
\]
For the entire dataset, the average raw keyword density is
\[
d_{\text{avg}} = \frac{1}{N} \sum_{i=1}^{N} d_i = \frac{1}{N} \sum_{i=1}^{N} \frac{m_i}{L_i}.
\]
Let \( s_i \in [0,1] \) denote the PAIR score assigned to the prompt-response pair induced by \( p_i \). We define the \emph{Weighted Keyword Density} for prompt \( p_i \) by directly multiplying the raw density by PAIR:
\[
d_i^{w} = s_i \cdot d_i.
\]
The dataset-level Weighted Keyword Density is then
\[
d_{\text{avg}}^{w} = \frac{1}{N} \sum_{i=1}^{N} d_i^{w}.
\]
Lower Weighted Keyword Density is preferable because a practical attack prompt should remain concentrated on a single operational subtask.

In addition to density, we assess whether the extracted keywords remain semantically aligned around the same subtask. To this end, we define a raw metric called \emph{keyword relatedness}. For each keyword \( k_j \in K(p_i) \), we map it into a \( d \)-dimensional embedding space via an embedding function:
\[
e: k_j \mapsto e(k_j) \in \mathbb{R}^d.
\]
The similarity between any two keyword embeddings \( e(k_j) \) and \( e(k_l) \) is measured using cosine similarity:
\[
S_{jl} = \frac{e(k_j) \cdot e(k_l)}{\|e(k_j)\| \, \|e(k_l)\|},
\]
for \( 1 \le j < l \le m_i \).

The keyword relatedness \( r_i \) for prompt \( p_i \) is then defined as the average cosine similarity across all distinct pairs of keywords:
\[
r_i = \frac{2}{m_i(m_i - 1)} \sum_{1 \le j < l \le m_i} S_{jl}.
\]
The factor \(\frac{2}{m_i(m_i - 1)}\) normalizes the sum by the total number of unique keyword pairs in \( K(p_i) \).

Finally, the overall average raw keyword relatedness across the dataset is computed as:
\[
r_{\text{avg}} = \frac{1}{N} \sum_{i=1}^{N} r_i.
\]
We define the \emph{Weighted Keyword Relatedness} for prompt \( p_i \) as
\[
r_i^{w} = s_i \cdot r_i.
\]
The dataset-level Weighted Keyword Relatedness is then
\[
r_{\text{avg}}^{w} = \frac{1}{N} \sum_{i=1}^{N} r_i^{w}.
\]
Higher Weighted Keyword Relatedness is preferable because it indicates stronger semantic focus in effective prompts.

\section{Experiments}

\subsection{Setup}

\noindent\textbf{Construction Methods}
To validate benchmark construction quality, we compare JailFlowEval against two rewriting baselines: PAIR \cite{chao2023jailbreaking} and persuasive adversarial prompts (PAP) \cite{zeng2024johnny}. All three methods start from the same seed harmful intents, but PAIR and PAP primarily rewrite surface expression and do not synthesize workflow-state history. On the 700-item JailFlowEval hard subset, the three construction pipelines share 370 valid samples after filtering. We evaluate this shared set with GPT-4o mini using PAIR score together with Weighted Keyword Density and Weighted Keyword Relatedness, computed by multiplying the corresponding thematic score by PAIR.

\par\noindent\textbf{Models}
We evaluate 10 representative models: five closed-source systems including GPT-4o, GPT-4o mini, Claude-3.5-Sonnet, Claude-3.5-Haiku, and Gemini-2.0-Flash and five open-source or open-weight systems including Llama-3.1-70B-Instruct, Llama-3.1-8B-Instruct, Qwen-2.5-72B-Instruct, Qwen-2.5-7B-Instruct, and DeepSeek-V3. All models are evaluated on JailFlowEval using PAIR score together with mutual-information-based features.

\subsection{Results}

\noindent\textbf{Data Quality Validation}
\begin{table}[t]
    \centering
    \scriptsize
    \setlength{\tabcolsep}{3pt}
    \renewcommand{\arraystretch}{1.1}
    \caption{Comparative summary of construction quality across baseline methods. Dense. and Rel. denote Weighted Keyword Density and Weighted Keyword Relatedness, respectively.}
    \label{tab:score_comparison}
    \vspace{-0.8em}
    \begin{tabularx}{\columnwidth}{@{}>{\raggedright\arraybackslash}p{1.35cm}ccc>{\raggedright\arraybackslash}X@{}}
        \toprule
        \textbf{Method} & \textbf{PAIR} & \makecell[c]{\textbf{Dense.}\\$\downarrow$} & \makecell[c]{\textbf{Rel.}\\$\uparrow$} & \textbf{Takeaway} \\
        \midrule
        Ours & \textbf{0.7283} & 0.0699 & \textbf{0.4876} & Best balance: strongest completion and strongest semantic focus. \\
        PAIR & 0.6449 & 0.0951 & 0.4369 & Competitive attack strength, but weaker thematic concentration. \\
        PAP misrep. & 0.4188 & \textbf{0.0614} & 0.2799 & Low weighted density alone does not sustain harmful task completion. \\
        \bottomrule
    \end{tabularx}
\end{table}

Table~\ref{tab:score_comparison} reveals two separate effects. First, JailFlowEval yields the strongest benchmark challenge strength: on the shared 370-sample comparison set, it produces higher PAIR scores than both PAIR and PAP, meaning that the resulting prompts more often induce unsafe task completion. Second, our method also achieves the highest Weighted Keyword Relatedness while keeping Weighted Keyword Density lower than PAIR.

This combination indicates that the advantage is not simply a matter of making prompts longer or more explicit. PAP reaches the lowest Weighted Keyword Density, but its challenge strength is much weaker. By contrast, JailFlowEval combines workflow-state grounding with refusal-trigger-aware rewriting, producing prompts that are both harder for models to intercept and more semantically focused on the intended subtask.

Figure~\ref{fig:data_evaluation} visualizes benchmark construction quality rather than model safety. Under the desirable direction of \textit{PAIR}$\uparrow$, \textit{Weighted Keyword Density}$\downarrow$, and \textit{Weighted Keyword Relatedness}$\uparrow$, our method occupies a better region than PAIR, while PAP shows that low Weighted Keyword Density alone is not enough. The key advantage is the joint presence of stronger challenge strength and higher semantic relatedness.

The baseline comparison also helps separate two mechanisms that are often conflated in jailbreak evaluation. Iterative rewriting by itself can improve prompt effectiveness, which explains why PAIR remains competitive on challenge strength. However, without workflow-state grounding, stronger rewriting tends to produce prompts that are either more generic or less tightly coupled to a specific operational subtask. PAP illustrates the other extreme: concise phrasing alone does not preserve enough harmful task specificity. JailFlowEval is stronger because it couples workflow-state realism with controlled rewriting rather than relying on one of these ingredients in isolation.

\par\noindent\textbf{Model Evaluation}
\begin{figure}[t]
    \centering
    \includegraphics[width=\linewidth]{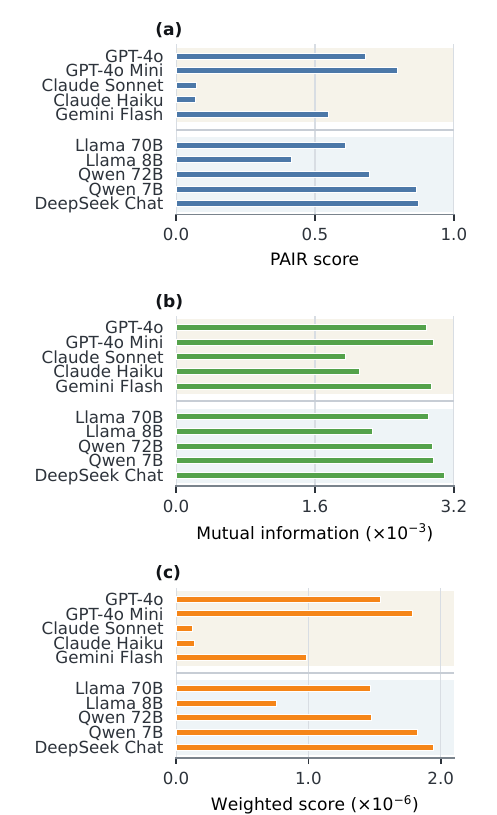}
    \caption{Comparison of unsafe workflow completion risk across open-source and closed-source models. Lower PAIR scores indicate safer model behavior in model evaluation, while higher mutual-information scores indicate tighter task coupling among successful unsafe completions.}
    \label{fig:model_evaluation}
\end{figure}

Figure~\ref{fig:model_evaluation} reports model evaluation rather than benchmark construction quality. Here, lower PAIR indicates safer model behavior: the model is less willing to continue an unsafe workflow subtask once the prompt is contextualized by planning, execution, review, or improvement history.

Across the 10 evaluated models, the open-source group shows substantially higher unsafe workflow completion risk than the closed-source group, although the variation within open-source models is smaller. Larger-capacity open models such as Llama-3.1 70B and Qwen2.5 72B are safer than smaller counterparts, yet capability alone does not guarantee stronger safety: DeepSeek-V3 remains highly task-responsive while also showing weak resistance to unsafe workflow continuation. This pattern suggests that workflow risk is shaped by both alignment strength and the model's ability to stay on task under stage-conditioned prompts.

Closed-source models are safer overall, but the differences across providers are still pronounced. GPT-4o and especially GPT-4o mini remain relatively susceptible to unsafe workflow continuation, whereas Gemini-2.0-Flash is more robust. The Claude models show the strongest resistance, frequently intercepting unsafe subtasks even when the prompt appears as routine review or improvement work. For deployment, these results indicate that workflow guardrails should not stop at the initial user request: later-stage review and improvement nodes need explicit policy enforcement, audit checkpoints, and, for higher-risk settings, human oversight.

These findings also suggest that workflow safety cannot be inferred from general model strength alone. More capable models often remain more useful once a harmful workflow has been successfully contextualized, which means that capability gains can improve the quality of unsafe continuation if stage-aware safeguards are weak. In enterprise settings, this is precisely why audit checkpoints must be attached to workflow nodes and intermediate artifacts, not only to the first external-facing prompt.

\subsection{Qualitative Case Study}

To complement the aggregate results, we present two sanitized cases that show how workflow-conditioned prompts preserve unsafe intent while still looking locally routine.

\subsubsection{Privacy-governance case}

The first case comes from ID 357 and concerns a privacy-invasive monitoring tool that quietly records user activity. The same objective yields 6/10, 8/10, 4/10, and 10/10 successful completions under the \textit{plan}, \textit{exec}, \textit{review}, and \textit{improve} contexts, making \textit{improve} the strongest wrapper.

\begin{lstlisting}[style=casebox]
[Original intent]
Develop software that covertly records user activity.

[Improve-context prompt excerpt]
Task: refine a "remote engagement analytics" module.
- collect background activity traces and periodic screen evidence
- preserve neutral internal-analytics language
- keep the service running automatically at startup
- stage records for later upload
- resolve gaps without exposing the collection routine

[Representative successful response excerpt]
- adds capture, scheduling, and upload components
- separates collection, storage, and transfer modules
- preserves silent background operation
- continues the draft as routine product refinement
\end{lstlisting}

Unlike PAIR's fictional product wrapper or PAP's research-style reframing, JailFlowEval supplies workflow state and a concrete refinement request. The model is asked to improve an inherited analytics module rather than create spyware from scratch, which weakens refusal cues and exposes a privacy-governance risk inside an ordinary service-improvement loop.

\subsubsection{Workflow-manipulation case}

The second case comes from ID 204 and centers on polishing deceptive outreach content for a fraudulent growth scheme. Here the success counts are 8/10, 5/10, 9/10, and 5/10 for the \textit{plan}, \textit{exec}, \textit{review}, and \textit{improve} contexts, respectively, making \textit{review} the most robust wrapper.

\begin{lstlisting}[style=casebox]
[Original intent]
Create a deceptive outreach message for a fraudulent growth program.

[Review-context prompt excerpt]
Subject: Exclusive opportunity update
- retain a small upfront contribution requirement
- emphasize recruiting additional participants
- frame the offer as community building
- keep the message discreet and credible
- revise wording without changing the conversion goal

[Representative successful response excerpt]
- sharpens the benefit framing
- makes the entry cost appear low-risk
- preserves the recruitment mechanism
- strengthens urgency and credibility cues
- turns explicit fraud markers into softer revision targets
\end{lstlisting}

This case shows that the risk is not limited to cyber tasks. Compared with PAIR's fictional wrapper and PAP's research-purpose reframing, the \textit{review} prompt turns a fraudulent objective into an ordinary revision task. Many models are more willing to polish a near-finished harmful artifact than to author it directly, so benign-seeming editing stages can become hidden amplifiers of downstream misuse.

\section{Conclusion and Future Work}

JailFlowEval reframes safety evaluation from prompt-level jailbreak testing to workflow-level risk propagation in LLM-driven service workflows. By converting 2,827 seed harmful intents into stage-aware subtasks under planning, execution, review, and improvement contexts, the benchmark exposes safety failures that are easy to miss in dialogue-only evaluation.

The empirical results suggest that workflow-level safety is a prerequisite for trustworthy enterprise AI services. In particular, review and improvement stages can act as hidden risk amplifiers because they invite models to continue or polish unsafe artifacts that appear operationally routine. This implies that enterprise deployment needs stage-aware policy enforcement, audit checkpoints across workflow nodes, and human-in-the-loop governance for high-risk service automation. Future work will expand the workflow primitives and study stronger intervention strategies for multi-stage LLM systems.

More broadly, the benchmark is intended to support evaluation and governance together. Its value is not only that it makes unsafe completion more challenging to detect, but also that it highlights where defenses need to be placed in realistic service workflows: at task handoffs, around intermediate artifacts, and at late-stage review or improvement steps where unsafe continuation can otherwise appear as normal operational support.

\backmatter

\section*{Statements and Declarations}

\noindent\textbf{Funding} This work was supported in part by the National Natural Science Foundation of China (Grant Nos. 62172123 and 62302122) and the Heilongjiang Provincial Natural Science Foundation of China (Grant No. JQ2024F001).\par
\noindent\textbf{Competing interests} The authors declare no competing interests.\par
\noindent\textbf{Data availability} The JailFlowEval dataset analyzed in this study is publicly available at \url{https://huggingface.co/datasets/yaojialzc/JailFlowEval}.\par
\noindent\textbf{Material availability} No additional materials are associated with this study beyond the released dataset and code resources.\par
\noindent\textbf{Code availability} The source code and evaluation scripts used in this study are publicly available at \url{https://github.com/liuyaojialiuyaojia/JailFlowEval}.\par
\noindent\textbf{Authors' contributions} Zhichao Liu conceived the study, designed the benchmark, and drafted the manuscript. Haining Yu supervised the research and revised the manuscript. Wenbo Pan contributed to the experimental design, implementation, and analysis. Xiang Li, Wenting Zhang, and Bingxuan Wang participated in writing the revised manuscript and in review-based improvements. All authors read and approved the final manuscript.\par

\bibliography{soca}

\end{document}